\pdfoutput=1
\documentclass[11pt]{article}

\usepackage[preprint]{acl}

\usepackage{times}
\usepackage{latexsym}

\usepackage[T1]{fontenc}

\usepackage[utf8]{inputenc}

\usepackage{microtype}

\usepackage{inconsolata}

\usepackage{graphicx}

\usepackage{amsmath}

\usepackage{booktabs}
\usepackage{multirow}

\usepackage{xcolor}
\usepackage{tikz}
\usepackage{mdframed}
\usepackage{fontawesome5}

\newcommand{\splitlabel}[3]{%
  \tikz[baseline=-0.6ex]{%
    \node[
      rounded corners=3pt,
      fill=#2,
      draw=#1,
      line width=0.8pt,
      inner xsep=4pt,
      inner ysep=2pt,
      font=\bfseries\small
    ] {#3};%
  }%
}

\definecolor{priorred}{RGB}{180, 50, 50}
\definecolor{priorfill}{RGB}{250, 220, 220}
\definecolor{visualblue}{RGB}{40, 110, 170}
\definecolor{visualfill}{RGB}{230, 240, 255}

\newcommand{\Prior}{\splitlabel{priorred}{priorfill}{Prior}}
\newcommand{\Visual}{\splitlabel{visualblue}{visualfill}{Visual}}

\title{Vision-Default, Prior-Override: Causal Mechanisms of Perception-Knowledge Conflict in Vision-Language Models}

\author{
  Niclas Lietzow\textsuperscript{1}, Danielle Bitterman\textsuperscript{2}, Carsten Eickhoff\textsuperscript{1},
  \\
  {\bf William Rudman}\textsuperscript{3}, {\bf Michal Golovanevsky}\textsuperscript{2} \\
  \textsuperscript{1}University of Tübingen \quad
  \textsuperscript{2}Harvard University \quad
  \textsuperscript{3}The University of Texas at Austin \\[0.4em]
  \small Correspondence: \texttt{niclas.lietzow@student.uni-tuebingen.de} \\
  \faGithub\ \small\url{https://github.com/nlietzow/vision-default-prior-override.git}
}
\begin{document}
\maketitle

\begin{abstract}
Vision-language models must reconcile visual evidence with memorized world knowledge when the two conflict. How they resolve this conflict shapes the reliability of multimodal systems, yet prior work characterizes it behaviorally without a component-level \textit{causal account}. We combine activation patching across three granularities (residual stream, attention heads, and MLP sublayers) with model-component ablation studies and mechanistic analysis. Across three VLM families, we find that visual grounding emerges by default, whereas prior grounding depends on a small set of causally necessary attention heads (2.5--4.8\%) concentrated in the second half of the network. These heads enable answers from stored world knowledge (e.g., ``red'' for a strawberry) despite conflicting visual input. Ablating them flips predictions from knowledge-grounded to visually grounded answers in 68--96\% of cases under prior-knowledge prompts, but changes only 0.8--7.5\% of visually grounded predictions, establishing an asymmetric causal structure. The identified heads decompose into \textit{routing} heads, which modulate information flow, and \textit{writing} heads, which directly project answer tokens into the residual stream. This structure is consistent across model families and scales, revealing a sparse causal circuit underlying perception-knowledge conflict in VLMs.
\end{abstract}

\section{Introduction}
\begin{figure}[t]
  \centering
  \includegraphics[width=0.9\columnwidth]{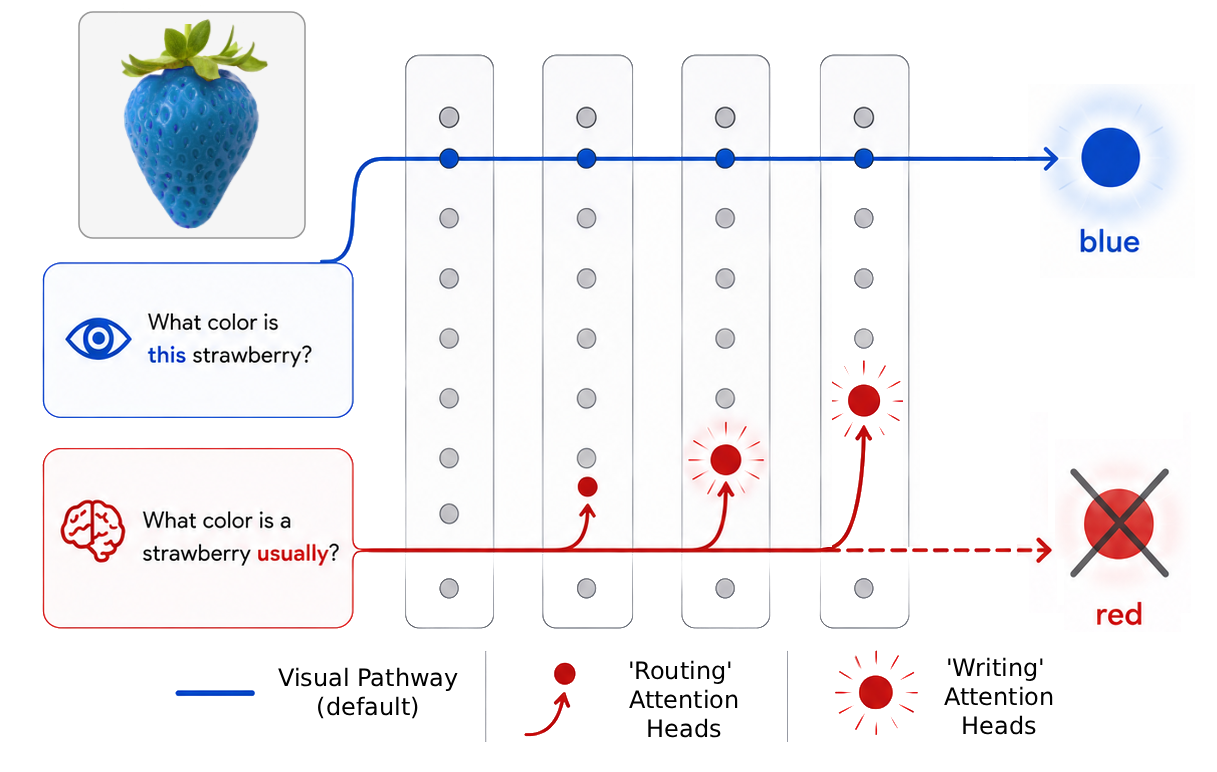}
  \caption{
Vision-language models resolve perception-knowledge conflict asymmetrically: visual grounding emerges by default, while prior knowledge depends on sparse late-layer routing and writing attention heads.
}
  \label{fig:overview}
\end{figure}

Recent work has increasingly questioned how Vision Language Models (VLMs) balance perceptual evidence with memorized world knowledge, especially when the two conflict \citep{golovanevsky-etal-2025-pixels, hua2025, ortu-etal-2026-seeing, zhang2025}.
For instance, when shown a visually conflicting image such as a blue strawberry and asked ``what color is this strawberry?'', VLMs often correctly report blue. Yet when asked ``what color is a strawberry usually?'', a question that should rely on prior knowledge rather than the image, the model frequently continues to respond based on the observed visual input \citep{golovanevsky-etal-2025-pixels}. This suggests that visual evidence can incorrectly override learned semantic knowledge even when the prompt requires a knowledge-grounded answer.
Understanding this interaction is important for improving the reliability of multimodal systems, particularly in determining when models should trust visual input versus retrieved world knowledge.

The perception-knowledge conflict has been studied from several complementary perspectives. Prior work has localized the conflict to mid-to-late network layers \citep{hua2025, golovanevsky-etal-2025-pixels}, identified candidate routing heads \citep{hua2025, ortu-etal-2026-seeing}, and shown that activation-level interventions can shift models between visual and prior grounding modes \citep{ortu-etal-2026-seeing, golovanevsky-etal-2025-pixels}. Other studies suggest that visual information degrades in late layers \citep{liu2025visual} while remaining partially recoverable from interpretable token representations \citep{neo2025towards}, and that modality selection under conflict follows predictable uncertainty dynamics \citep{zhang2025}. These findings characterize \emph{where} the conflict emerges and show that the behavior \emph{can be controlled}, but they do not explain the underlying mechanism by which VLMs \textbf{resolve conflicts} between visual evidence and stored knowledge.

\begin{enumerate}
\item We provide the first component-level causal account of perception-knowledge conflict in VLMs, identifying specific attention heads and MLP sublayers that mediate the decision. These components decompose into early routing heads that modulate information flow and late writing heads that directly project answer tokens into the residual stream.

\item We establish an asymmetric causal structure across all evaluated models: visual grounding surfaces by default, while prior grounding depends on active injection by a sparse 2.5--4.8\% of attention heads concentrated in the second half of the network. Ablating these heads flips 68--96\% of conflict predictions under prior grounding, but only 0.8--7.5\% under visual grounding. MLP sublayers show the same directional asymmetry at substantially weaker magnitudes, consistent with an amplifier rather than a primary routing role.

\item We show that this routing-and-writing circuit generalizes across three architecturally distinct VLM families and multiple model scales (Qwen-VL 3B/7B, LLaVA-NeXT 7B, PaliGemma 3B/10B). However, the routing implementation diverges across architectures: Qwen-VL and LLaVA-NeXT redistribute attention between image and text tokens, whereas PaliGemma routes through differences in the attended representations. These results reveal a shared causal architecture governing how VLMs resolve conflicts between what they see and what they know.
\end{enumerate}

\section{Related Work}

\paragraph{Perception-Knowledge Conflict in VLMs.}
Recent work has studied the perception-knowledge conflict in VLMs from behavioral, representational, and intervention-based perspectives. Multiple studies showed that VLMs frequently override memorized world knowledge with conflicting visual evidence under counterfactual inputs \citep{golovanevsky-etal-2025-pixels, ortu-etal-2026-seeing}. Other work localized the conflict to mid-to-late network layers \citep{hua2025, golovanevsky-etal-2025-pixels}, where visual representations remain partially interpretable even as visual information degrades in later layers \citep{liu2025visual}. Correlational analyses identified candidate routing heads in image-versus-caption \citep{hua2025} and factual-versus-counterfactual \citep{ortu-etal-2026-seeing} settings, while activation-level interventions demonstrated that model behavior can be shifted between \protect\Visual{} grounding (following visual evidence) and \protect\Prior{} grounding (following stored world knowledge) \citep{golovanevsky-etal-2025-pixels}. Uncertainty-based frameworks further predict which modality dominates under conflict \citep{zhang2025}.

However, existing approaches remain largely behavioral or correlational. Steering-vector methods manipulate activations without identifying the responsible circuitry, while correlational analyses cannot establish causal necessity.

\citet{nooralahzadeh2026arbitrationfailureperceptualblindness} moves toward a causal analysis through full-sequence residual-stream patching, but studies the complementary setting in which models answer from prior knowledge despite a visual prompt. We instead study the more common regime in which visual evidence overrides prior knowledge, while also operating at finer granularity through per-head and per-MLP interventions. Unlike the relatively balanced trade-offs observed for parametric-versus-retrieval conflict in unimodal language models \citep{jin-etal-2024-cutting}, we find a strongly asymmetric structure in VLMs, where visual grounding surfaces by default and prior knowledge requires active override. Our setting is the multimodal analogue of context-memory conflict in language models \citep{xu-etal-2024-knowledge-conflicts}, where entity substitutions in the context reveal over-reliance on parametric memory \citep{longpre2021entitybased}; here the counterfactual image supplies the conflicting contextual evidence.

\paragraph{Mechanistic Interpretability of VLMs.}
Mechanistic interpretability in VLMs builds on frameworks developed for language models, including causal mediation analysis for identifying sparse mediators of model behavior \citep{vig2020}, causal tracing for localizing factual associations \citep{NEURIPS2022_6f1d43d5}, and logit lens, which projects intermediate representations into vocabulary space to inspect what token-level information they encode \citep{nostalgebraist2020}. Methodological work on activation patching has further emphasized in-distribution corruption and logit-difference metrics as reliable choices for causal intervention studies \citep{zhang2024towards}.

\citet{palit2023towards} introduced activation patching for the text decoder of VLMs, and \citet{golovanevsky-etal-2025-vlms} generalized activation patching to causal interventions over both text and image representations, finding attention heads whose importance is shared across corrupted modalities. \citet{jiang2025interpreting} further extended logit-lens-style analysis to VLM image tokens. Other mechanistic analyses identified sparse task-specific attention heads in vision transformers \citep{10.1007/978-3-031-72775-7_15}, localized factual retrieval in multimodal models through causal tracing \citep{NEURIPS2024_0dfe31d6}, analyzed head importance under semantic image edits \citep{wang-etal-2025-v}, and linked compositional failures in CLIP vision encoders to neuron-level superposition in MLP layers \citep{11375415}.

\paragraph{Multimodal Information Routing.}
Recent work has studied how VLMs route information between visual and textual modalities, including which layers process image tokens \citep{neo2025towards}, where text-copying resides \cite{rudman-etal-2026-mechanisms}, how cross-modal attention patterns develop \citep{11094641}, and where modality-specific circuits diverge \citep{nikankin2025same}. \citet{liu2025visual} found that mid-layer image value tokens encode sufficient information for perception tasks, but that visual information degrades in later layers, where input-agnostic key tokens actively suppress perception. These findings suggest that late-layer components play a central role in multimodal arbitration and information routing.

Our results connect these representational findings to the underlying routing mechanism, identifying the sparse components that determine whether a VLM follows visual evidence or prior knowledge under conflict.



\section{Methods}

\subsection{Task Setting}
\label{sec:task-setup}
We use the Visual-Counterfact dataset \citep{golovanevsky-etal-2025-pixels}, which contains 469 counterfactual color images: everyday objects  recolored to conflict with world knowledge (e.g., a blue strawberry, an orange elephant), each paired with a color-identification question. Two image variants exist: the \emph{original} (real-world colors) and the \emph{counterfactual} (recolored). We evaluate the model under two grounding modes: \protect\Visual{} (``What color is this \{object\} here?''), which prompts the model to report what it sees, and \protect\Prior{} (``What color is \{a(n) object\} usually?''), which asks for memorized world knowledge. We refer to a forward pass under the \protect\Visual{} prompt as \emph{visual-grounded} and under the \protect\Prior{} prompt as \emph{prior-grounded}. The conflict condition is \protect\Prior{} grounding on the counterfactual image: the model sees a blue strawberry but is asked what color a strawberry usually is. Here, visual evidence and memorized knowledge produce competing answers (see Table~\ref{tab:inference} for example prompts and image variants).

We evaluate five model sizes spanning three architecturally distinct VLM families: Qwen-VL-2.5 (3B, 7B) \citep{bai2025qwen25vltechnicalreport}, LLaVA-NeXT 7B with a Mistral backbone \citep{llava_next}, and PaliGemma (3B, 10B) with a Gemma 2 backbone \citep{paligemma2}. Parameter counts range from 3B to 10B. These models are the standard testbed for prior interpretability work on counterfactual and conflict-resolution VLM settings \citep{golovanevsky-etal-2025-pixels, golovanevsky-etal-2025-vlms, ortu-etal-2026-seeing, hua2025}, enabling direct comparison with the existing literature. All models are accessed via NNsight \citep{fiotto-kaufman2025nnsight}, which provides activation-level read and write access during the forward pass without modifying model code.

All quantitative analyses are restricted to \emph{correctly conflicting} examples: those where the unmodified model produces the counterfactual color under \protect\Visual{} grounding and the original color under \protect\Prior{} grounding on the counterfactual image, each matching the expected answer (see Table~\ref{tab:correctly_conflicting}). This ensures that every measurement reflects a genuine conflict resolution rather than noise from examples where the model already fails at one or both grounding modes.

\subsection{Activation Patching}
We perform activation patching at the last token position (the position where the model generates its answer) in both swap directions (P2V and V2P, defined below), to identify which components carry information that causally determines how the conflict is resolved. Prior work identifies this position as the most salient site for patching instruction-tuned models~\citep{golovanevsky-etal-2025-vlms, NEURIPS2025_085713-3551}.

\paragraph{Notation.} For each counterfactual image, let $x_V$ and $x_P$ denote the model inputs under \protect\Visual{} and \protect\Prior{} grounding (same image, different prompt), and let $\text{logit}_t(x)$ denote the model's last-token logit for token $t$ on a clean forward pass on input $x$. We write $a_c(x)$ for the activation at component $c$ at the last token position during that clean forward pass, and define the patched-run logit
\begin{equation}
\tilde{\ell}^{\,c}_t(x \mid x') \;=\; \text{logit}_t(x)\,\big|_{a_c \,\leftarrow\, a_c(x')},
\end{equation}
the last-token logit on input $x$ when the activation at $c$ has been replaced by the value cached from a clean forward pass on $x'$.

For each dataset example, we run the model under both grounding modes on the same counterfactual image, caching intermediate activations at the target component. Because the image is held fixed and only the prompt varies, this contrast targets components that mediate which information source the model surfaces. We then patch one grounding's cached activation into the other grounding's forward pass and measure the effect on the prediction. This yields two patching directions:
\begin{itemize}
\item \textbf{P2V} (Prior $\to$ Visual; target $x_V$, source $x_P$): Patch a prior-grounded component activation into the visual-grounded forward pass. If the prediction shifts toward the prior answer, that component carries prior-relevant information.
\item \textbf{V2P} (Visual $\to$ Prior; target $x_P$, source $x_V$): Patch a visual-grounded activation into the prior-grounded forward pass, testing whether the component carries visual-relevant information.
\end{itemize}

We patch at three granularities, each evaluated at the last token position:
\begin{equation}
a_c \in \big\{\, r^{(\ell)},\; z_h^{(\ell)},\; m^{(\ell)} \,\big\}
\end{equation}
where $r^{(\ell)}$ is the \emph{residual-stream} output at layer $\ell$ (the full layer output combining attention, MLP, and residual connections), $z_h^{(\ell)}$ is the output vector of \emph{attention head} $h$ at the $W_O$ input at layer $\ell$, and $m^{(\ell)}$ is the \emph{MLP} sublayer output at layer $\ell$.

\paragraph{Restoration score.} For a component $c$ patched in direction $d \in \{\text{P2V}, \text{V2P}\}$, we define:
\begin{equation}
R_d(c) = \frac{\Delta_d^{\text{patched}}(c) - \Delta_d^{\text{target}}}{\Delta_d^{\text{source}} - \Delta_d^{\text{target}}}
\end{equation}
where $\Delta = \text{logit}(t_{\text{orig}}) - \text{logit}(t_{\text{cf}})$ is the logit difference between the original-color and counterfactual-color answer tokens, and the superscripts denote the clean source run, the clean target run, and the target run after patching component $c$, respectively. A restoration score of 1 means the patch fully restores the source run's logit difference; 0 means no effect.

\paragraph{Flip rate.}
As a discrete complement to $R_d(c)$, we report the fraction of examples where patching changes the argmax prediction. For each example $i$ in direction $d$, let $\hat{t}_i^{\,c} := \arg\max_t \tilde{\ell}^{\,c}_t(x_T^{(i)} \mid x_S^{(i)})$ denote the patched argmax and $t_i^* := \arg\max_t \text{logit}_t(x_T^{(i)})$ the clean argmax. The flip rate over $N$ examples is:
\begin{equation}
F_d(c) = \frac{1}{N} \sum_{i=1}^{N} \mathbf{1}\!\left[\hat{t}_i^{\,c} \neq t_i^*\right].
\end{equation}
In words, $F_d(c)$ is the share of examples in which intervention on component $c$ changes the model's top answer. Concretely, $F_{\text{V2P}}(c) = 0.33$ means that for 33\% of examples, patching $c$'s visual-grounded activation into the prior-grounded run flips the top prediction from the prior-consistent answer (e.g., \emph{red} for a strawberry) to the visual-consistent answer (e.g., \emph{blue}).

\paragraph{Component classification.}
For each component, we compute the average restoration vector $(R_{\text{P2V}}, R_{\text{V2P}})$ across examples. PCA is then applied across all components, and those beyond $\pm 2\sigma$ on the first principal component are classified as \emph{promoting} (high restoration in both directions) or \emph{suppressing} (negative restoration opposing the patched direction). The procedure is applied separately to attention heads and MLP sublayers.

\subsection{Causal Component Ablation}
Activation patching tests sufficiency: can a component restore a behavior? Component ablation tests necessity: does removing the component remove the behavior?

We zero-ablate the target head or MLP output at the last token position during the forward pass. We test both \emph{individual} ablations (one component at a time) and \emph{group} ablations (all promoting or suppressing components simultaneously). Comparing the two reveals whether the effect is concentrated in single components or distributed across the group.

As with all analyses (Section~\ref{sec:task-setup}), ablations are restricted to correctly conflicting examples. We report the flip rate: the fraction of examples where the prediction changes from the expected answer to the competing answer under each grounding mode.

\subsection{Mechanistic Characterization}

Patching and model component ablation establish \emph{which} components matter; mechanistic analysis characterizes \emph{how} they achieve their causal effects. We perform two complementary analyses on all classified heads.

\paragraph{Attention pattern analysis.} For each classified head, we extract attention weights at the last token position under both grounding modes on the counterfactual image. We aggregate attention over all image token positions into a single \emph{image-attention fraction} and compute the delta (visual $-$ prior grounding). A positive delta indicates greater attention to image tokens under visual grounding and a shift toward text tokens under prior grounding, consistent with dynamic attention routing.

\paragraph{Logit lens on head-output differences.} We extract each classified head's output vector at the last token position (before recombination via $W_O$) under both grounding modes, compute the difference vector $\mathbf{h}_{\text{visual}} - \mathbf{h}_{\text{prior}}$, and project it through the output projection matrix $W_O$ and the model's unembedding matrix into vocabulary logit space. We then check whether the original-color and counterfactual-color answer tokens appear in the top-$k$ or bottom-$k$ positions ($k{=}20$) of the resulting distribution. A high hit rate indicates the head directly encodes the answer token difference into the residual stream.


\section{Results}

\subsection{Behavioral Evidence of Visual Override}

\begin{table}[t]
\centering
\small
\setlength{\tabcolsep}{3.5pt}
\begin{tabular}{lcccccc}
\toprule
& \multicolumn{3}{c}{\protect\Visual{}} & \multicolumn{3}{c}{\protect\Prior{}} \\
\cmidrule(lr){2-4} \cmidrule(lr){5-7}
Model & Orig & CF & $\Delta$ & Orig & CF & $\Delta$ \\
\midrule
Qwen-VL 3B    & 92.1 & 91.9 & \textbf{0.2} & 95.1 & 17.7 & \textbf{77.4} \\
Qwen-VL 7B    & 93.4 & 86.1 & \textbf{7.3} & 95.7 & 55.7 & \textbf{40.0} \\
LLaVA-NeXT 7B & 91.9 & 86.4 & \textbf{5.5} & 94.7 & 21.7 & \textbf{73.0} \\
PaliGemma 3B  & 92.3 & 88.1 & \textbf{4.2} & 91.7 & 32.4 & \textbf{59.3} \\
PaliGemma 10B & 91.7 & 87.0 & \textbf{4.7} & 93.0 & 44.6 & \textbf{48.4} \\
\bottomrule
\end{tabular}
\caption{Accuracy (\%) across grounding modes and image variants. \protect\Visual{} asks ``What color is this strawberry?''; \protect\Prior{} asks ``What color is a strawberry usually?''. Orig = real-color image; CF = counterfactual; $\Delta$ = Orig $-$ CF. Visual grounding is robust ($\Delta \leq 7.3$), while Prior collapses under conflict ($\Delta = 40.0$--$77.4$).}
\label{tab:inference}
\end{table}

Table~\ref{tab:inference} summarizes inference accuracy across all conditions. When no conflict exists (\protect\Visual{} with either image variant, and \protect\Prior{} on the original image), all five models achieve 86--96\% accuracy regardless of architecture or scale. The conflict condition (\protect\Prior{} grounding with counterfactual image) produces a dramatic collapse: accuracy drops to 17.7--55.7\%, as models report what they see rather than what they know. Larger models resist visual override more effectively (Qwen-VL 3B to 7B improves from 17.7\% to 55.7\%; PaliGemma 3B to 10B improves from 32.4\% to 44.6\%), but no model eliminates it. At matched 7B scale, Qwen-VL achieves 55.7\% accuracy under conflict, while LLaVA-NeXT reaches only 21.7\%. This replicates the general phenomenon observed by \citet{golovanevsky-etal-2025-pixels}, while showing that differences in conflict behavior cannot be explained by model scale alone, motivating the need for cross-architecture causal analysis.

\subsection{The Decision Forms in a Critical Window}
\label{sec:critical-window}

\begin{figure}[t]
  \centering
  \includegraphics[width=\columnwidth]{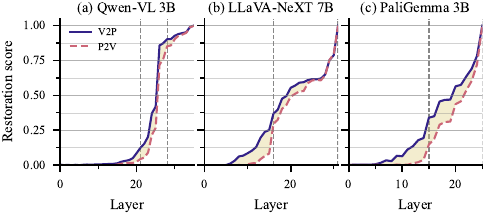}
  \caption{Residual stream restoration scores $R_d(\ell)$ by layer for three representative models. P2V (dashed) and V2P (solid) patching directions are shown; the shaded region highlights the V2P--P2V asymmetry, and vertical dashed lines mark the critical window boundaries. Across models, V2P restoration rises earlier and more strongly than P2V, indicating that visual information is established before prior knowledge. Different architectures exhibit distinct transition dynamics, ranging from sharp late-layer shifts to gradual multi-layer accumulation. See Appendix~\ref{app:res_stream}, Figure~\ref{fig:res_stream_all} for all five models.}
  \label{fig:res_stream}
\end{figure}

We first ask where in the network the conflict is resolved. Residual-stream patching localizes this to a critical window in the second half of the network. Letting $\Delta R_d(\ell) = R_d(\ell) - R_d(\ell-1)$ denote the per-layer increment in restoration score for direction $d$ (with $R_d(0) = 0$), we define the critical window for each direction $d$ as the smallest consecutive layer range $[\ell_a, \ell_b]$ for which $\sum_{\ell=\ell_a}^{\ell_b} \Delta R_d(\ell) \geq 0.8 \sum_\ell \Delta R_d(\ell)$. Across all five models, these windows span 7--16 layers and begin at 52--76\% of network depth. Although broad at the layer level, the underlying circuit becomes sparse at the component level (Section~\ref{sec:heads}).

Within this window, V2P restoration (patching visual information into the prior-grounded run) consistently reaches high flip rates before P2V restoration (patching prior information into the visual-grounded run). The gap in layer where each direction first reaches 50\% flip rate ($F_d = 0.5$) ranges from 3 layers in Qwen-VL 7B to 19 in LLaVA-NeXT 7B (Appendix~\ref{app:res_stream}, Table~\ref{tab:flip50}). This ordering holds despite distinct architectural dynamics (Figure~\ref{fig:res_stream}; additional models in Appendix~\ref{app:res_stream}, Figure~\ref{fig:res_stream_all}): Qwen-VL shows a sharp late rise, LLaVA-NeXT plateaus before a final-layer jump, and PaliGemma accumulates gradually across many layers.

This pattern provides the first evidence of the \emph{vision-default, prior-override} mechanism: visual information appears earlier in the residual stream, while prior knowledge emerges later. If both grounding modes were processed symmetrically, V2P and P2V restoration would accumulate at similar rates. Instead, the consistent lag of P2V suggests that visual grounding is the default pathway, while prior knowledge requires additional computation, which we test directly through component-level patching and ablation.

\subsection{A Sparse Set of Heads Drives the Decision}
\label{sec:heads}

\begin{figure}[t]
  \centering
  \includegraphics[width=\columnwidth]{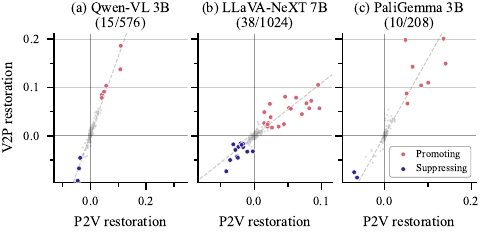}
  \caption{Attention head classification by patching restoration score. Each point is one head, positioned by mean P2V and V2P restoration. Promoting heads are shown in red and suppressing heads in blue; the dashed line indicates the first principal component. Across all models, most heads cluster near zero, with only a sparse subset (2.5--4.8\%) strongly mediating the conflict. See Appendix~\ref{app:head_classification}, Figure~\ref{fig:head_classification_all} for all five models.}
  \label{fig:head_classification}
\end{figure}

Having localized the decision to a critical layer window, we next ask which specific attention heads carry the causal signal. The \emph{vision-default, prior-override} hypothesis predicts that this set should be sparse: if visual grounding is the default pathway, only a small minority of heads should actively inject prior knowledge. Figure~\ref{fig:head_classification} plots each head by its mean P2V and V2P restoration scores $(R_{\text{P2V}}, R_{\text{V2P}})$ averaged over correctly conflicting examples. Most heads cluster near the origin, indicating a negligible causal effect.

To identify the sparse subset that departs from this baseline, we project each head's restoration vector onto the first principal component (PC1) of the joint distribution and classify heads beyond $\pm 2\sigma$. Heads with positive PC1 projections are labeled \emph{promoting}: patching them restores the source-grounding answer in both directions, indicating that they actively mediate the routing decision. Heads with negative projections are labeled \emph{suppressing}: patching them pushes predictions away from the source answer, indicating opposition to the patched direction. In Figure~\ref{fig:head_classification}, promoting heads appear in red and suppressing heads in blue.

The classification recovers the predicted sparse circuit across all models: only 2.5--4.8\% of heads are classified. These heads concentrate primarily in the second half of the network, overlapping with the critical window identified by residual-stream patching. The residual-stream asymmetry also sharpens at head level: promoting heads produce strong V2P flip rates but minimal P2V effects, consistent with visual grounding as the default pathway and prior grounding as an active override. By contrast, MLP effects are weaker and substantially less consistent across architectures, suggesting that attention heads dominate the routing mechanism. The selected set is stable under resampling: re-classifying on four fifths of the examples returns the full-data promoting set exactly in 19 of 25 model-fold runs, with the few heads that move sitting at the classification threshold (Appendix~\ref{app:kfold}).

\subsection{Vision as the Default Pathway}
\label{sec:knockout}

\begin{figure}[t]
  \centering
  \includegraphics[width=\columnwidth]{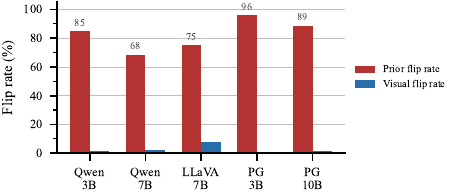}
  \caption{Flip rates under promoting-head group ablation. Qwen = Qwen-VL; LLaVA = LLaVA-NeXT; PG = PaliGemma. Dark bars show prior-grounding flips; light bars show visual-grounding flips. Ablating promoting heads disrupts prior grounding (68--96\%) while leaving visual grounding largely intact (0.8--7.5\%), consistent with vision as the default pathway.}
  \label{fig:knockout}
\end{figure}


\begin{table}[t]
\centering
\small
\setlength{\tabcolsep}{4pt}
\begin{tabular}{@{}lcccc@{}}
\toprule
& \multicolumn{2}{c}{Attention} & \multicolumn{2}{c}{MLP} \\
\cmidrule(lr){2-3} \cmidrule(lr){4-5}
Model & Prom. & Supp. & Prom. & Supp. \\
\midrule
Qwen-VL 3B
  & 84.9 / 1.4 & 19.2 / 0.0 & 74.0 / 1.4 & 11.0 / 0.0 \\
Qwen-VL 7B
  & 68.4 / 2.4 & 0.5 / 0.9  & 28.8 / 1.4 & -- \\
LLaVA 7B
  & 75.0 / 7.5 & 6.2 / 0.0  & --          & 23.8 / 1.2 \\
PG 3B
  & 95.9 / 0.8 & 5.8 / 0.0  & --          & 18.2 / 0.0 \\
PG 10B
  & 88.7 / 1.7 & 6.2 / 0.0  & --          & 11.3 / 1.1 \\
\bottomrule
\end{tabular}
\caption{Group ablation flip rates (Prior / Visual). Promoting attention-head ablations consistently flip prior-grounded predictions (68--96\%) while minimally affecting visual grounding (0.8--7.5\%). MLP effects are weaker but directionally similar. ``--'' indicates that no classified MLPs were available for ablation.}
\label{tab:knockout}
\end{table}

The activation patching results show that a sparse set of heads carries information relevant to the conflict resolution. Model-component ablation now tests the complementary question: are these heads \emph{necessary}? The answer establishes the central finding of this paper.

Ablating all promoting attention heads flips prior-grounded predictions in 68--96\% of correctly conflicting examples across all five models, while changing visual-grounded predictions in only 0.8--7.5\% (Table~\ref{tab:knockout}, Figure~\ref{fig:knockout}). Thus, removing these heads largely eliminates prior grounding while leaving visual grounding intact. The same heads are both sufficient to restore prior grounding when patched in (per-head V2P flip rates 26.9--74.0\% versus P2V 0.0--2.6\%) and necessary to sustain it under ablation, establishing visual grounding as the default pathway and prior grounding as an active override.

Suppressing-head group ablations produce weaker effects, with prior grounding flip rates of 0.5--19.2\%. Their removal does not substantially affect either grounding mode, indicating that they are not necessary for routing.

MLP group ablations show the same directional asymmetry as attention heads (11--74\% prior flips versus 0--1.4\% visual flips), but at substantially weaker magnitudes. Qwen-VL 3B exhibits the strongest MLP effects, reaching 74.0\% prior flip rate compared to 84.9\% for attention heads, while PaliGemma and LLaVA-NeXT contain no promoting MLP layers and reach at most 23.8\% prior flip rate through suppressing-layer ablations. These results suggest that MLPs amplify memorized prior knowledge once a routing pathway is selected, rather than serving as primary routing components, consistent with prior work identifying MLP sublayers as the main site of factual knowledge storage in transformers \citep{geva-etal-2021-transformer, geva-etal-2022-transformer, dai-etal-2022-knowledge, NEURIPS2022_6f1d43d5}. Larger models further weaken MLP effects, reinforcing the amplifier interpretation.


One remaining question is whether the observed asymmetry reflects the prompt-based contrast used throughout the paper, which varies the grounding prompt while holding the image fixed and therefore does not directly isolate circuits that read visual color information. As a robustness check, we repeat the analysis with a complementary visual-circuit contrast that varies the image while holding the prompt fixed. Under this contrast, ablating promoting attention heads flips visual-grounded predictions in 1.8--40.6\% of examples, still well below the 68--96\% prior-grounding flips produced by the primary analysis. Both contrasts, therefore, support the same conclusion: visual grounding is more robust than prior grounding.


\subsection{Routing and Writing Attention Heads}

\begin{figure*}[t]
  \centering
  \includegraphics[width=0.48\linewidth]{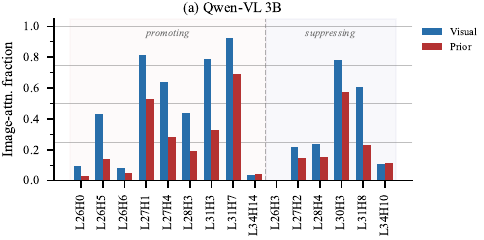} \hfill
  \includegraphics[width=0.48\linewidth]{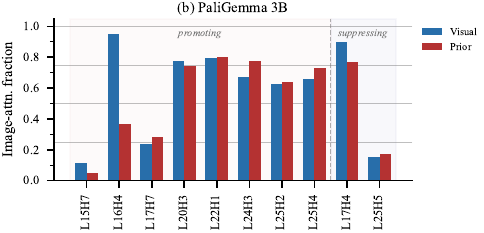}
  \caption{Image-attention fraction for classified heads under \protect\Visual{} (blue) and \protect\Prior{} (red) grounding. Qwen-VL 3B (a) shifts attention between image and text tokens depending on the grounding mode (mean delta $+0.22$), while PaliGemma 3B (b) maintains high image-attention under both modes with near-zero delta ($+0.05$). These patterns illustrate two routing regimes: attention-routing (Qwen-VL, LLaVA-NeXT) and representation-routing (PaliGemma). See Appendix~\ref{app:attention_logit_lens}, Figure~\ref{fig:attention_routing_all} for all five models.}
  \label{fig:attention_routing}
\end{figure*}

\paragraph{Attention routing.} Two architecturally distinct routing mechanisms emerge from the attention analysis (Figure~\ref{fig:attention_routing}). In Qwen-VL and LLaVA-NeXT, classified heads dynamically shift attention between image and text tokens depending on the grounding mode, with mean image-attention deltas (visual $-$ prior) of $+0.19$ to $+0.24$. Under prior grounding, these heads redirect attention away from image tokens toward the textual context. In PaliGemma, classified heads maintain high image-attention (0.5--0.9) under both conditions, with near-zero mean deltas. Although equally causally important, these heads operate through changes in the attended representations rather than through attention redistribution. Both promoting and suppressing heads follow the same architecture-specific pattern.

\paragraph{Routing versus writing heads.}
Projecting head-output differences into vocabulary space reveals which classified heads directly encode the answer token. In every model, a small set of late-layer heads places the counterfactual color among the top-20 predicted tokens in over 80\% of examples, while most other classified heads show near-zero hit rates despite strong causal effects. Two functional roles therefore emerge: early \emph{routing heads}, which redirect information flow, and late \emph{writing heads}, which project the final decision into vocabulary space. This routing-then-writing decomposition holds consistently across architectures despite differences in the routing mechanism itself. The dissociation also constrains how restoration should be read: if classified heads simply wrote answer tokens, causal importance and direct-write rate would coincide, yet most classified heads remain necessary under ablation while showing near-zero hit rates.


\subsection{Generalization Beyond Color Conflicts}
\label{sec:whoops}

\begin{table}[t]
\centering
\small
\setlength{\tabcolsep}{4pt}
\begin{tabular}{@{}lccccc@{}}
\toprule
& & \multicolumn{2}{c}{Recovered} & Ablation & \\
\cmidrule(lr){3-4} \cmidrule(lr){5-5}
Model & $n$ & Prom. & Supp. & Prior / Vis. & $p$ \\
\midrule
Qwen-VL 3B  & 26 & 8/9   & 6/6  & 57.7 / 11.5 & .008 \\
Qwen-VL 7B  & 33 & 19/21 & 7/8  & 48.5 / 6.1  & .001 \\
LLaVA 7B    & 21 & 20/24 & 9/14 & 38.1 / 14.3 & .227 \\
PG 3B       & 17 & 7/8   & 1/2  & 58.8 / 11.8 & .039 \\
PG 10B      & 27 & 12/12 & 3/5  & 59.3 / 3.7  & {<}.001 \\
\bottomrule
\end{tabular}
\caption{Generalization to WHOOPS!. \emph{Recovered}: heads of the Visual-Counterfact circuit re-discovered on WHOOPS!. \emph{Ablation}: prior- and visual-grounding flip rates when the promoting group is ablated, with a two-sided McNemar exact test on the paired per-example outcomes. The asymmetry holds in all five models and is significant in four.}
\label{tab:whoops}
\end{table}

Every result so far uses Visual-Counterfact, where the conflict is between two colors and the answer space is binary. To test whether the identified circuit generalizes, we repeat the full pipeline on counterfactual images from the WHOOPS! benchmark \citep{10377616}, re-captioned from the WHOOPS-AHA! curation of \citet{ortu-etal-2026-seeing} to fit our \protect\Visual{} and \protect\Prior{} prompt template. The re-captioned items span several conflict categories, from counterfactual object shapes (e.g., a cube-shaped watermelon) to unnatural environments (e.g., a snowman in the desert).  Answers are drawn from an open vocabulary rather than a fixed color pair: for an image of a snowplow clearing sand, the two prompts elicit \emph{snow} and \emph{sand}, which take the role of the original and counterfactual colors. As before we retain only conflicting examples, here those whose top-1 prediction differs between the two prompts, leaving $n = 17$ to $33$ per model (Appendix~\ref{app:whoops}).

\paragraph{The circuit is re-discovered.} Re-running the classification procedure from scratch on WHOOPS! recovers nearly all of the promoting heads originally selected on Visual-Counterfact: 8 of 9 in Qwen-VL 3B, 19 of 21 in Qwen-VL 7B, 20 of 24 in LLaVA-NeXT, 7 of 8 in PaliGemma 3B, and all 12 in PaliGemma 10B (Table~\ref{tab:whoops}). Attention heads that merely write color directions into the residual stream could not mediate the conflicts in WHOOPS!, where answers are not strictly colors. Thus, the promoting heads we identify are not color-specific circuit components. Recovery of the suppressing set is generally weaker and more variable (from 1 of 2 to 6 of 6), consistent with the limited causal role those heads play throughout.

\paragraph{The asymmetry persists.} Ablating the promoting group identified on WHOOPS! flips prior-grounded predictions in 38.1--59.3\% of examples while changing visual-grounded predictions in only 3.7--14.3\%. The direction of the asymmetry holds in all five models, and a McNemar exact test on the paired per-example outcomes reaches significance in four of them ($p \leq .039$); LLaVA-NeXT is the exception ($p = .227$) on the smallest conflicting subset ($n = 21$). Effect sizes are smaller compared to Visual-Counterfact, where the same ablation flips 68--96\% of prior-grounded predictions against 0.8--7.5\% of visual-grounded ones, as expected on a harder task whose answers are open-vocabulary and often multi-token. Suppressing-group ablation is close to null throughout (prior flips 3.0--19.2\%, significant in no model), as on the color task. The vision-default, prior-override mechanism is therefore not an artifact of the binary color setting.

\subsection{Cross-Architecture Generalization}

The central claim of \emph{vision-default, prior-override} holds across all three VLM families and five model sizes. Six properties generalize consistently:
\begin{enumerate}
\itemsep0pt
\item \textbf{Ablation asymmetry:} prior flip 68--96\%, visual flip 0.8--7.5\%.
\item \textbf{Critical window} in the second half of the network (52--100\% depth).
\item \textbf{V2P precedes P2V} by 3--19 layers.
\item \textbf{Both head types present,} with 2.5--4.8\% of heads classified.
\item \textbf{Two-stage routing$\to$writing mechanism} (0\%$\to${>}80\% logit lens hit rate).
\item \textbf{MLP asymmetry} in the same direction as heads but 1.2--8$\times$ weaker.
\end{enumerate}

Four properties are architecture-specific: the routing mechanism (attention redistribution in Qwen-VL and LLaVA-NeXT, mean image-attention delta $+0.19$ to $+0.24$, against near-zero deltas in PaliGemma), the accumulation dynamics, the presence of promoting MLP layers in Qwen-VL alone, and the degree of circuit redundancy. Despite these implementation differences the core asymmetry remains consistent, suggesting a convergent computational strategy.

\section{Discussion}
Our results suggest that perception-knowledge conflict in VLMs is not primarily a failure of perception or missing world knowledge. Models often perceive the counterfactual image correctly and retain the relevant semantic knowledge, yet still default to the visual input even when the task requires prior knowledge instead. This has important implications for reliability. Improving perception alone will not fix cases where models must ignore misleading visual evidence, and improving factual knowledge alone will not ensure that the knowledge is used.

The sparse prior-grounding circuit we identify gives this asymmetry a concrete mechanistic basis. Prior knowledge depends on a small set of attention heads, whereas visual grounding remains robust when those heads are removed. This helps explain why conflicting visual evidence is so difficult for VLMs to ignore.

Although the mechanism generalizes across architectures, the implementation differs substantially. Qwen-VL and LLaVA-NeXT reroute attention between image and text tokens, while PaliGemma changes the representations extracted from attended tokens without strongly changing the attention pattern itself. The same behavioral asymmetry therefore emerges from different internal computations, suggesting that future control methods may need to be architecture-specific.

This account also gives existing control methods a target. Activation-level steering \citep{golovanevsky-etal-2025-pixels} and attention redistribution \citep{ortu-etal-2026-seeing} shift behavior in this conflict without explaining why the two directions are not equally easy to steer. The asymmetry we identify in this study supplies one: prior-grounded answers require engaging the sparse override circuit these heads form, while visually grounded answers work with the default pathway. Language models show comparable receptiveness to coherent conflicting evidence \citep{xie2024adaptivechameleon}.

\section{Conclusion}

VLMs resolve perception-knowledge conflicts through an asymmetric mechanism in which visual grounding surfaces by default while prior knowledge requires active injection by a sparse set of causally necessary attention heads. These heads, only 2.5--4.8\% of the total and concentrated late in the network, decompose into early routing heads that modulate information flow and late writing heads that encode answer tokens directly. The mechanism holds across three architecturally distinct VLM families, and the circuit is recovered on open-vocabulary conflicts beyond color. The identified heads are principled targets for controlling when a VLM follows what it sees.

\section*{Limitations}
\paragraph{Conflict types.} Our main experiments use color-property conflicts from Visual-Counterfact, which provides a controlled and interpretable setting for isolating mechanisms of visual-textual conflict resolution. Section~\ref{sec:whoops} extends the analysis to open-vocabulary conflicts from WHOOPS!, spanning unnatural environments, principle conflicts, incorrect usage, object shape, and color inversion, and the same circuit is recovered there. 

\paragraph{Prompt contrast.} Both datasets vary the grounding prompt while holding the image fixed, so the measured asymmetry may partly reflect how the two prompts are routed rather than arbitration between perception and memory alone. Appendix~\ref{app:visual_circuit} varies the image while holding the prompt fixed and recovers a strongly overlapping head set, and WHOOPS! reproduces the mechanism on a second dataset, though its contrast is prompt-based as well. A full $2\times2$ design crossing prompt type with image type, testing whether the classified heads show a super-additive effect in conflict cases beyond additive prompt-only and image-only effects, would isolate the arbitration component more directly. We see this as the natural next step for this line of work.

\paragraph{Model scale.} We evaluate models in the 3B--10B parameter range, following scales commonly used in prior mechanistic interpretability work \cite{golovanevsky-etal-2025-pixels, hua2025, ortu-etal-2026-seeing} and enabling tractable intervention-based analysis. We prioritize architectural diversity over scale. Larger models may develop different strategies as their capacity and memorized knowledge increase.

\paragraph{Intervention position.} Our interventions target the last token position, where the model produces its answer, consistent with standard practice in mechanistic interpretability studies of autoregressive models \cite{golovanevsky-etal-2025-vlms, NEURIPS2025_085713-3551}. Residual-stream patching at that position restores the grounding decision within a late critical window, so the decision-relevant information does flow through it, but our claims concern this late-stage arbitration only. Components contributing earlier in the sequence, for example during the processing of image tokens, are outside the scope of the analysis and their role may be underestimated.



\bibliography{custom}

\appendix

\section{Dataset and Example Selection}
\label{app:dataset}

The Visual-Counterfact dataset \citep{golovanevsky-etal-2025-pixels} contains 469 examples of common objects with digitally recolored images. Each example pairs an object (e.g., banana, elephant) with its original color (e.g., yellow, gray) and a counterfactual color (e.g., blue, orange). Two examples (radish and spider) were excluded due to overlapping original and counterfactual colors, leaving 467 examples for analysis.

All quantitative analyses are restricted to correctly conflicting examples as defined in Section~\ref{sec:task-setup}. Table~\ref{tab:correctly_conflicting} reports the number per model.

\begin{table}[t]
\centering
\small
\begin{tabular}{@{}lrr@{}}
\toprule
Model & Total & Correct conflict \\
\midrule
Qwen-VL 3B  & 467 &  73 \\
Qwen-VL 7B  & 467 & 212 \\
LLaVA-NeXT 7B   & 467 &  80 \\
PaliGemma 3B  & 467 & 121 \\
PaliGemma 10B & 467 & 177 \\
\bottomrule
\end{tabular}
\caption{Number of correctly conflicting examples per model, used for all quantitative analyses.}
\label{tab:correctly_conflicting}
\end{table}

\section{Full Residual Stream Curves}
\label{app:res_stream}

\begin{figure*}[t]
  \centering
  \includegraphics[width=\linewidth]{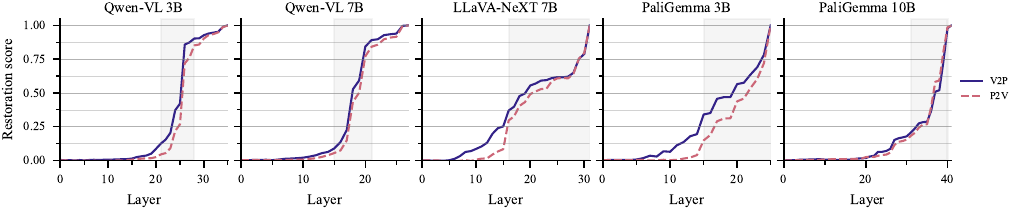}
  \caption{Residual stream restoration scores for all five models. P2V (dashed) and V2P (solid) patching directions. The main paper (Figure~\ref{fig:res_stream}) shows three representative models; this figure includes Qwen-VL 7B and PaliGemma 10B.}
  \label{fig:res_stream_all}
\end{figure*}

Figure~\ref{fig:res_stream_all} shows residual stream patching restoration scores for all five models. The main paper (Figure~\ref{fig:res_stream}) shows three representative models; this figure adds Qwen-VL 7B and PaliGemma 10B.

\begin{table}[t]
\centering
\begin{tabular}{lccc}
\toprule
Model & V2P & P2V & Gap \\
\midrule
Qwen-VL 3B   & 24 & 28 &  4 \\
Qwen-VL 7B   & 18 & 21 &  3 \\
LLaVA-NeXT 7B    & 12 & 31 & 19 \\
PaliGemma 3B   & 13 & 25 & 12 \\
PaliGemma 10B  & 31 & 40 &  9 \\
\bottomrule
\end{tabular}
\caption{Layer at which V2P and P2V patching directions first reach 50\% flip rate (Gap = P2V $-$ V2P). V2P consistently precedes P2V, with gaps ranging from 3 to 19 layers.}
\label{tab:flip50}
\end{table}

\section{Head Classification Details}
\label{app:head_classification}

\begin{figure*}[t]
  \centering
  \includegraphics[width=\linewidth]{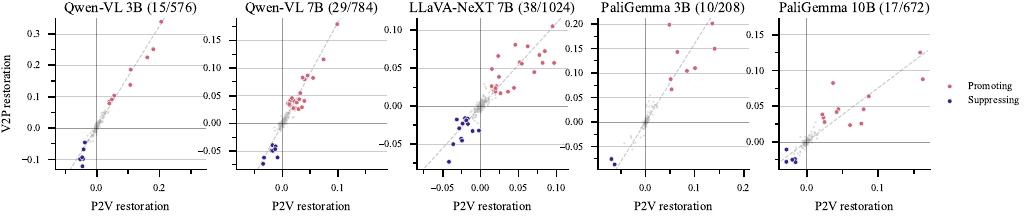}
  \caption{Attention head classification scatters for all five models; the dashed line shows the PC1 axis used for the $\pm 2\sigma$ classification. The main paper (Figure~\ref{fig:head_classification}) shows three representative models; this figure adds Qwen-VL 7B and PaliGemma 10B.}
  \label{fig:head_classification_all}
\end{figure*}

Figure~\ref{fig:head_classification_all} shows classification scatters for all five models. Tables~\ref{tab:promoting_heads} and~\ref{tab:suppressing_heads} list all classified heads by model.

\begin{table*}[t]
\centering
\small
\begin{tabular}{@{}llp{12cm}@{}}
\toprule
Model & Count & Promoting heads \\
\midrule
Qwen-VL 3B & 9 & L26H0, L26H5, L26H6, L27H1, L27H4, L28H3, L31H3, L31H7, L34H14 \\
Qwen-VL 7B & 21 & L17H21, L18H8, L18H9, L18H24, L19H22, L19H23, L19H24, L20H1, L20H3, L20H5, L21H5, L21H19, L22H1, L22H13, L23H6, L23H11, L24H21, L24H27, L26H24, L26H25, L26H26 \\
LLaVA 7B & 24 & L13H27, L15H7, L16H0, L16H1, L16H10, L17H0, L17H5, L17H21, L18H8, L18H10, L18H12, L19H9, L20H21, L22H20, L24H21, L24H22, L28H13, L29H2, L29H12, L29H14, L30H29, L31H22, L31H25, L31H27 \\
PG 3B & 8 & L15H7, L16H4, L17H7, L20H3, L22H1, L24H3, L25H2, L25H4 \\
PG 10B & 12 & L27H15, L32H11, L33H4, L36H1, L36H2, L37H12, L39H0, L39H2, L39H7, L39H13, L40H6, L40H10 \\
\bottomrule
\end{tabular}
\caption{All promoting attention heads (PCA $> +2\sigma$) by model. LLaVA = LLaVA-NeXT; PG = PaliGemma.}
\label{tab:promoting_heads}
\end{table*}

\begin{table*}[t]
\centering
\small
\begin{tabular}{@{}llp{12cm}@{}}
\toprule
Model & Count & Suppressing heads \\
\midrule
Qwen-VL 3B & 6 & L26H3, L27H2, L28H4, L30H3, L31H8, L34H10 \\
Qwen-VL 7B & 8 & L18H7, L20H2, L20H21, L21H14, L22H25, L25H25, L26H22, L27H3 \\
LLaVA 7B & 14 & L14H14, L16H2, L16H3, L17H24, L18H9, L18H11, L18H28, L21H6, L22H29, L24H20, L28H14, L29H0, L29H13, L31H20 \\
PG 3B & 2 & L17H4, L25H5 \\
PG 10B & 5 & L30H8, L31H7, L32H10, L40H7, L40H11 \\
\bottomrule
\end{tabular}
\caption{All suppressing attention heads (PCA $< -2\sigma$) by model. LLaVA = LLaVA-NeXT; PG = PaliGemma.}
\label{tab:suppressing_heads}
\end{table*}

\section{Individual Knockout Ablation Results}
\label{app:individual_knockout}

Table~\ref{tab:individual_knockout} shows individual head ablation flip rates for the two leading promoting heads in each model.

\begin{table}[t]
\centering
\begin{tabular}{llr}
\toprule
\textbf{Model} & \textbf{Head} & \textbf{Prior flip (\%)} \\
\midrule
Qwen-VL 3B & L26H5 & 41.1 \\
Qwen-VL 3B & L31H3 & 37.0 \\
Qwen-VL 7B & L20H5 & 9.4 \\
Qwen-VL 7B & L23H6 & 7.1 \\
LLaVA-NeXT 7B & L31H27 & 26.2 \\
LLaVA-NeXT 7B & L31H22 & 22.5 \\
PaliGemma 3B & L15H7 & 58.7 \\
PaliGemma 3B & L20H3 & 36.4 \\
PaliGemma 10B & L40H6 & 18.1 \\
PaliGemma 10B & L37H12 & 10.2 \\
\bottomrule
\end{tabular}
\caption{Individual head ablation flip rates for top promoting heads. PaliGemma 3B shows concentrated effects (L15H7 alone at 58.7\%), while Qwen-VL 7B is distributed (maximum 9.4\%).}
\label{tab:individual_knockout}
\end{table}

\section{Compensation and Redundancy Analysis}
\label{app:compensation}

To quantify redundancy among the promoting heads, we ask how often a single-head ablation flips the prediction on examples that the group ablation flips. In Qwen-VL 7B, 85.5\% of examples flipped by the group ablation cannot be flipped by any single head: the effect is distributed across 21 promoting heads, each making partial contributions that are individually insufficient but collectively necessary. In PaliGemma 3B, the effect is far more concentrated: only 35.3\% of group-flipped examples are fully compensated, and a single head (L15H7) alone accounts for 60.3\% of the group effect. Even when the effect is distributed, individual head ablations still reduce the prior answer's logit margin by +0.2 to +1.7 points, confirming that each head makes a genuine partial contribution even when that contribution is insufficient to flip the prediction on its own. The rate of redundancy is thus architecture-dependent, not a uniform property of the mechanism.

\section{Visual-Circuit Ablation (Robustness Check)}
\label{app:visual_circuit}

As a robustness check on the central asymmetry reported in Section~\ref{sec:knockout}, we run an alternative ablation experiment using a visual-circuit contrast: hold the prompt fixed while varying the image between original and counterfactual variants. This contrast targets components that read color from the image rather than components that mediate the prompt-driven choice. We re-run the full PCA classification on the resulting restoration scores and ablate the classified groups. Results are shown in Table~\ref{tab:knockout_visual_circuit}.

\begin{figure*}[t]
  \centering
  \includegraphics[width=\linewidth]{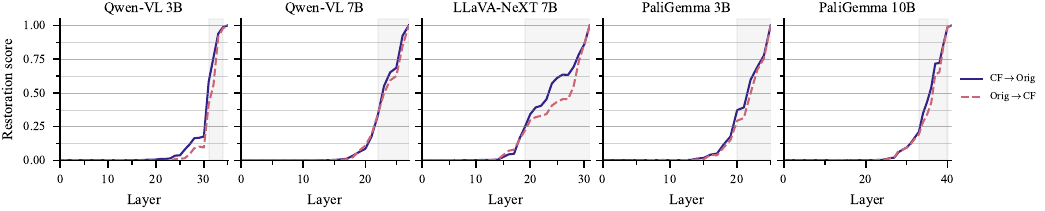}
  \caption{Residual-stream restoration scores under the visual-circuit contrast (vary image, hold prompt) for all five models. Dashed: patching original-image activations into the counterfactual-image forward pass. Solid: patching counterfactual-image activations into the original-image forward pass. Shaded region marks the critical window (smallest range covering $>$80\% of restoration). The window locations (7--16 layer span, starting at 52--76\% of network depth) match the primary contrast's localization (Figure~\ref{fig:res_stream_all}).}
  \label{fig:res_stream_visual_circuit_all}
\end{figure*}

Residual-stream patching under the visual-circuit contrast localizes restoration to a critical window in the second half of each model, spanning 7--16 layers and starting at 52--76\% of network depth, matching the primary contrast's localization (Section~\ref{sec:critical-window}, Figure~\ref{fig:res_stream_visual_circuit_all}). MLP patching under this contrast produces sparse, low-magnitude effects, mirroring the MLP-as-amplifier pattern from the primary analysis. PCA classification on the visual-circuit attention-head restoration scores identifies a comparably sparse set (1.6--5.8\% of heads per model, versus 2.5--4.8\% under the primary contrast), with 61\% of visual-circuit-classified heads also classified under the primary contrast, further evidence that the two contrasts engage overlapping rather than separate components.

\begin{table}[t]
\centering
\small
\begin{tabular}{@{}llrr@{}}
\toprule
& & \multicolumn{2}{c}{Flip rate (\%)} \\
\cmidrule(l){3-4}
Model & Ablation & Prior & Visual \\
\midrule
\multirow{4}{*}{Qwen-VL 3B}
  & Attn prom.  & 1.4  & 40.6 \\
  & Attn supp.  & 0.0  & 0.0 \\
  & MLP prom.   & 71.0 & 0.0 \\
  & MLP supp.   & 10.1 & 0.0 \\
\midrule
\multirow{4}{*}{Qwen-VL 7B}
  & Attn prom.  & 0.5  & 9.0 \\
  & Attn supp.  & 0.5  & 0.0 \\
  & MLP prom.   & 28.5 & 1.5 \\
  & MLP supp.   & --   & --  \\
\midrule
\multirow{4}{*}{LLaVA 7B}
  & Attn prom.  & 3.8  & 32.1 \\
  & Attn supp.  & 9.0  & 0.0 \\
  & MLP prom.   & --   & --  \\
  & MLP supp.   & 38.5 & 1.3 \\
\midrule
\multirow{4}{*}{PG 3B}
  & Attn prom.  & 35.7 & 17.4 \\
  & Attn supp.  & 6.1  & 0.0 \\
  & MLP prom.   & --   & --  \\
  & MLP supp.   & 16.5 & 0.0 \\
\midrule
\multirow{4}{*}{PG 10B}
  & Attn prom.  & 25.7 & 1.8 \\
  & Attn supp.  & 1.2  & 0.0 \\
  & MLP prom.   & --   & --  \\
  & MLP supp.   & 10.5 & 1.2 \\
\bottomrule
\end{tabular}
\caption{Group ablation flip rates for visual-circuit-classified components across five VLMs. Visual flip rates (1.8--40.6\%) stay well below the 68--96\% prior-flip rate produced by the primary prompt-contrast ablations (Table~\ref{tab:knockout}). PaliGemma models show high prior flip rates under attention-head promoting ablation. Dashes indicate no components classified in that category. LLaVA = LLaVA-NeXT; PG = PaliGemma.}
\label{tab:knockout_visual_circuit}
\end{table}

\section{MLP Analysis Details}
\label{app:mlp}

\begin{figure*}[t]
  \centering
  \includegraphics[width=\linewidth]{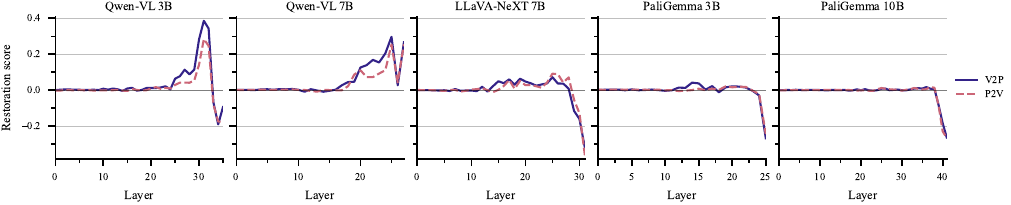}
  \caption{MLP restoration scores across layers for all five models. Effects are sparse, with only a few layers showing moderate contributions.}
  \label{fig:mlp_restoration}
\end{figure*}

MLP patching restoration scores are sparse across all models (Figure~\ref{fig:mlp_restoration}). Only Qwen-VL models have classified promoting MLP layers (L30--32 in Qwen-VL 3B; L25, L27 in Qwen-VL 7B). LLaVA-NeXT and PaliGemma models have only suppressing classifications or near-zero effects.

PaliGemma 10B shows near-zero MLP restoration scores across all layers ($\leq 0.02$), suggesting minimal MLP involvement in the conflict decision for this architecture. Late-layer MLPs in several models produce negative restoration scores, suggesting active suppression of the patched direction.

\section{Attention Patterns and Logit Lens}
\label{app:attention_logit_lens}

\begin{figure*}[t]
  \centering
  \includegraphics[width=\linewidth]{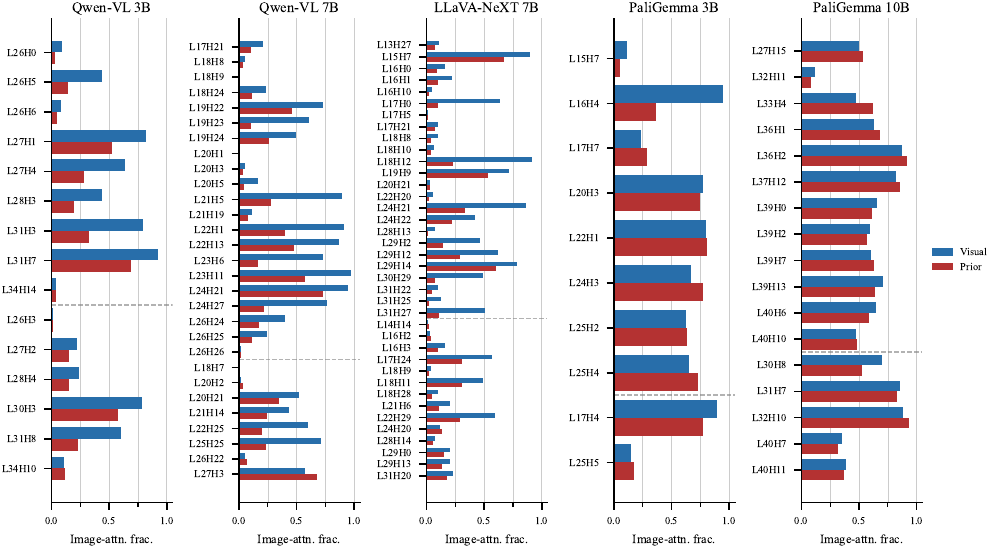}
  \caption{Image-attention fraction for all classified heads across five models, under \protect\Visual{} (blue) and \protect\Prior{} (red) grounding. Promoting and suppressing heads are separated by the dashed line within each panel. Qwen-VL and LLaVA-NeXT show large visual$-$prior gaps (attention routing); PaliGemma maintains high image-attention under both conditions (representation-routing).}
  \label{fig:attention_routing_all}
\end{figure*}

\begin{figure*}[t]
  \centering
  \includegraphics[width=\linewidth]{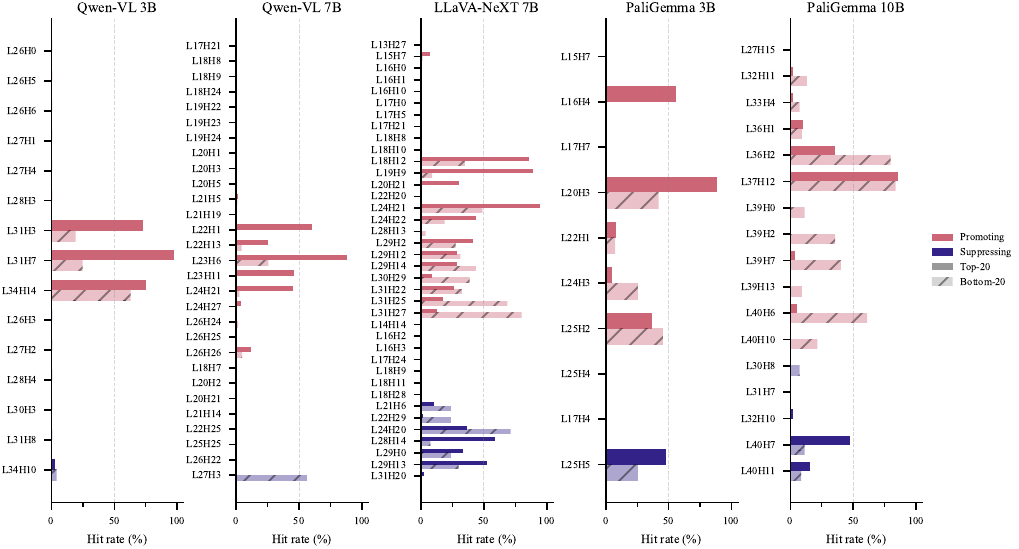}
  \caption{Logit-lens hit rates on head-output differences for all classified heads across five models. Top-20 hit rate indicates how often the counterfactual color appears among the 20 highest-ranked tokens in the projected head output difference. Late-layer heads consistently show high hit rates ($>$80\%), while earlier classified heads show 0\%.}
  \label{fig:logit_lens_all}
\end{figure*}

Figure~\ref{fig:attention_routing_all} reports per-head image-attention fractions under both grounding modes for all classified heads across the five models, extending the two-model view in the main paper (Figure~\ref{fig:attention_routing}). Figure~\ref{fig:logit_lens_all} reports logit-lens top-20 hit rates for all classified heads across the five models.

\section{WHOOPS! Generalization Details}
\label{app:whoops}

\paragraph{Dataset construction.} The WHOOPS! benchmark \citep{10377616} contains synthetic images that deliberately violate commonsense expectations. We use the subset curated by \citet{ortu-etal-2026-seeing} as WHOOPS-AHA!, re-captioned into our two-prompt template so that each item yields a \protect\Prior{} question (``\dots{} generally?'') and a \protect\Visual{} question (``\dots{} here?'') over the same image. The 42 re-captioned items span unnatural environments, principle conflicts, incorrect usage, object shape, and color inversion. Patching at the last token position requires the two prompts of a pair to tokenize to the same length. This was verified offline under the Qwen tokenizer for all 42 pairs, and every patching runner additionally asserts input-shape equality per example at run time, so the property holds under each model's own tokenizer for every example entering the reported results.

\paragraph{Answer tokens and example selection.} Answer tokens are the model's own top-1 predictions rather than a hand-specified answer key: the prior answer is the argmax under the \protect\Prior{} prompt and the visual answer the argmax under the \protect\Visual{} prompt. As in the main experiments we keep only conflicting examples, here those where the two predictions differ. The resulting subsets are model-specific and range from 17 to 33 of the 42 items (Table~\ref{tab:whoops_res_stream}). All 42 re-captioned pairs were inspected by hand to confirm that the two answers are semantically meaningful, so every per-model conflicting subset is covered. One LLaVA-NeXT example produced ablation outputs but no patching outputs; we exclude it throughout, so that classification and ablation draw on the same examples within each model. Every runner is otherwise byte-identical to the one used for Visual-Counterfact; only the question, image, answer tokens, and output directory differ.

\begin{table}[t]
\centering
\small
\setlength{\tabcolsep}{2.5pt}
\begin{tabular}{@{}lcccccc@{}}
\toprule
Model & $n$ & $L$ & P2V & V2P & Depth (\%) & V2P/P2V \\
\midrule
Qwen-VL 3B & 26 & 36 & 23--28 & 20--27 & 63--80 & 25 / 26 \\
Qwen-VL 7B & 33 & 28 & 16--22 & 14--21 & 52--78 & 18 / 20 \\
LLaVA 7B   & 21 & 32 & 12--29 & 13--31 & 35--94 & 16 / 16 \\
PG 3B      & 17 & 26 & 17--25 & 16--25 & 68--100 & 16 / 24 \\
PG 10B     & 27 & 42 & 26--39 & 27--40 & 66--98 & 31 / 38 \\
\bottomrule
\end{tabular}
\caption{Residual stream patching on WHOOPS!. $L$ is the number of layers; P2V and V2P give the smallest consecutive window covering more than 80\% of incremental restoration in each direction; \emph{Depth} is that window for the averaged curve as a fraction of network depth; the last column gives the first layer at which V2P and P2V respectively reach a 50\% flip rate.}
\label{tab:whoops_res_stream}
\end{table}

\paragraph{Residual stream.} The critical window falls in the second half of the network in four of the five models (Table~\ref{tab:whoops_res_stream}). LLaVA-NeXT is the exception, with a wide window spanning 35 to 94\% of depth, consistent with the plateau-then-jump accumulation it shows on Visual-Counterfact. V2P reaches a 50\% flip rate before P2V in four of the five models, by 1 to 8 layers, and simultaneously in LLaVA-NeXT. The Qwen-VL 3B window (layers 22 to 28, 63 to 80\% depth) matches the one measured on Visual-Counterfact (layers 21 to 28) except for its first layer.

\begin{table}[t]
\centering
\small
\setlength{\tabcolsep}{4pt}
\begin{tabular}{@{}lccc@{}}
\toprule
& & \multicolumn{2}{c}{Promoting / suppressing} \\
\cmidrule(lr){3-4}
Model & Heads & Visual-Counterfact & WHOOPS! \\
\midrule
Qwen-VL 3B & 576  & 9/6 (2.6\%)  & 13/9 (3.8\%) \\
Qwen-VL 7B & 784  & 21/8 (3.7\%) & 23/9 (4.1\%) \\
LLaVA 7B   & 1024 & 24/14 (3.7\%) & 26/18 (4.3\%) \\
PG 3B      & 208  & 8/2 (4.8\%)  & 7/1 (3.8\%) \\
PG 10B     & 672  & 12/5 (2.5\%) & 14/5 (2.8\%) \\
\bottomrule
\end{tabular}
\caption{Attention heads classified by the same PCA procedure at $\pm 2\sigma$, run independently on each dataset. Percentages are of all heads in the model. The proportion classified is comparable across datasets, so the recovery rates in Table~\ref{tab:whoops} are not an artifact of WHOOPS! selecting a much larger head set.}
\label{tab:whoops_classification}
\end{table}

\paragraph{Head classification.} Running the classification independently on each dataset selects a comparable fraction of heads (2.5--4.8\% on Visual-Counterfact, 2.8--4.3\% on WHOOPS!; Table~\ref{tab:whoops_classification}), so the recovery rates reported in Section~\ref{sec:whoops} do not follow from WHOOPS! simply classifying more heads. The promoting heads not re-identified on WHOOPS! are few and scattered: L27H4 in Qwen-VL 3B; L22H1 and L23H11 in Qwen-VL 7B; L17H5, L19H9, L22H20 and L28H13 in LLaVA-NeXT; L16H4 in PaliGemma 3B; and none in PaliGemma 10B.

\paragraph{MLP sublayers.} MLP contributions are sparse and late in both Qwen models, spanning layers 25 to 35 and 18 to 27 with peak V2P restoration of 0.34 and 0.29. They are close to absent in PaliGemma, where at most three layers exceed a mean absolute restoration of 0.05 and peak V2P restoration is 0.04 or below. Final-layer MLP restoration is negative in four of the five models ($-0.10$ to $-0.29$), matching the late-layer negativity seen on Visual-Counterfact; Qwen-VL 7B is the exception at $+0.24$. As on the color task, MLP effects are weaker than attention head effects and point in the same direction.

\section{Head Selection Stability}
\label{app:kfold}

Section~\ref{sec:heads} selects the classified heads and Section~\ref{sec:knockout} evaluates them by ablation. This appendix checks whether the selected set depends on the particular examples used to discover it.

\begin{table}[t]
\centering
\small
\setlength{\tabcolsep}{4pt}
\begin{tabular}{@{}lcc@{}}
\toprule
Model & Exact & Jaccard \\
\midrule
Qwen-VL 3B & 4/5 & 0.90--1.00 \\
Qwen-VL 7B & 4/5 & 0.95--1.00 \\
LLaVA 7B   & 1/5 & 0.88--1.00 \\
PG 3B      & 5/5 & 1.00 \\
PG 10B     & 5/5 & 1.00 \\
\bottomrule
\end{tabular}
\caption{Five-fold stability of the head classification. \emph{Exact} counts folds in which re-classifying on four fifths of the examples returns exactly the full-data promoting set; the last column gives the range of Jaccard overlap with the full-data promoting set across folds. Seed 42.}
\label{tab:kfold}
\end{table}

\paragraph{$K$-fold stability.} We partition each model's classification examples into five folds, re-run the PCA classification at $\pm 2\sigma$ on the remaining four fifths, and compare the selected set against the full-data classification (Table~\ref{tab:kfold}). The promoting set is recovered exactly in 19 of 25 model-fold runs, and Jaccard overlap with the full-data promoting set never falls below 0.88. Deviating folds differ by one to three heads. Every head that moves across folds lies between 1.9$\sigma$ and 2.4$\sigma$ on the classification axis, close to the $\pm 2\sigma$ threshold, so the instability is confined to borderline cases rather than to the core of the circuit. LLaVA-NeXT shows the most movement across folds (1 of 5 exact), consistent with it having the largest classified set.

Taken together with the independent re-discovery of the circuit on WHOOPS! (Section~\ref{sec:whoops}), which is a prediction on a disjoint dataset rather than a resampling of the same one, head selection is robust to which examples are used to discover it.

\end{document}